\documentclass{article} %
\usepackage{iclr2027_conference,times}

\usepackage[utf8]{inputenc}
\usepackage[T1]{fontenc}

\usepackage{amsmath,amsfonts,bm}

\def\eqref#1{equation~\ref{#1}}

\def\1{\bm{1}}

\DeclareMathAlphabet{\mathsfit}{\encodingdefault}{\sfdefault}{m}{sl}
\SetMathAlphabet{\mathsfit}{bold}{\encodingdefault}{\sfdefault}{bx}{n}

\usepackage{booktabs}
\usepackage{array,longtable}
\usepackage{nicefrac}
\usepackage{microtype}
\usepackage{xcolor}
\usepackage{amssymb}
\usepackage{mathtools}
\usepackage{amsthm}
\usepackage{graphicx}
\usepackage{adjustbox}
\usepackage{multirow}
\usepackage{colortbl}
\usepackage{arydshln}
\usepackage{comment}
\usepackage{tabularx}
\usepackage{caption}
\usepackage{subcaption}
\usepackage{hyperref}
\usepackage{url}
\usepackage[capitalize,noabbrev]{cleveref}

\newcommand{\modelname}{Quicksviewer}

\definecolor{mypink}{RGB}{239,43,159}

\title{Quicksviewer: An LMM with Early Video Compression using Annealed Gumbel-Softmax}

\author{%
Ji Qi$^{\diamond}$, Yuan Yao$^{\dagger}$\thanks{Corresponding author: \texttt{yaoyuanthu@gmail.com}.}\hspace{0.4em}, Yushi Bai$^{\diamond}$, Bin Xu$^{\diamond}$, Juanzi Li$^{\diamond}$, Zhiyuan Liu$^{\diamond}$, Tat-Seng Chua$^{\dagger}$ \\
$^{\diamond}$Tsinghua University \quad $^{\dagger}$National University of Singapore \\[1pt]
\texttt{qijithu@gmail.com, yaoyuanthu@gmail.com}
}

\iclrfinalcopy %
\begin{document}

\maketitle
\lhead{}

\vspace{-20pt}
\begin{center}
  \centering
  \includegraphics[width=\textwidth]{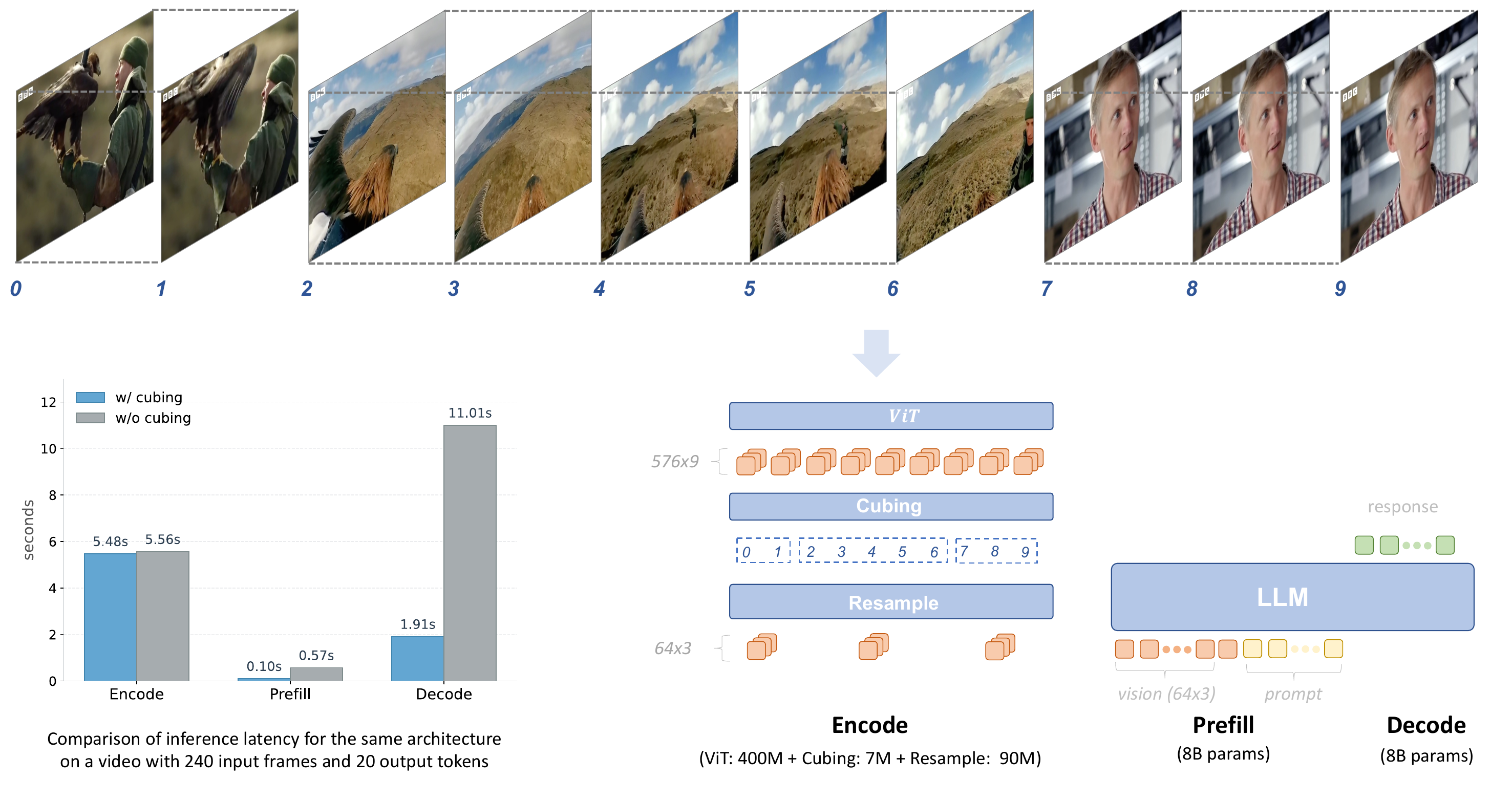}
  \vspace{-5pt}
  \captionof{figure}{\textbf{Left:} With the same architecture and input/output lengths, early video compression via cubing substantially reduces LLM prefill and decoding latency. \textbf{Right:} A lightweight cubing network partitions inherently nonuniform video frames (events of 2, 5, and 3 frames) into variable-length cubes, followed by a 3D resampler with spatiotemporal awareness, enabling efficient LLM processing.}
  \label{fig:teaser_fig}
\end{center}

\begin{abstract}
Large Multimodal Models (LMMs) uniformly perceive video frames, creating computational inefficiency for videos with inherently varying temporal redundancy.
This paper present \textbf{Quicksviewer}, an LMM with new perceiving paradigm that partitions a nonuniform redundant video into varying cubes using Gumbel Softmax, followed by a unified resampling for each cube to achieve efficient video understanding.
We further propose a Gumbel-noise annealing strategy that enables stable end-to-end LMM training with the discrete cubing network.
This simple and intuitive approach dynamically compress video online based on the momentum of frame differences, significantly reducing spatiotemporal complexity (overall 45$\times$ compression rate), while enabling efficient training with large receptive field.
We train the model from a language backbone through three progressive stages, each incorporating lengthy videos on average of 420s/1fps thanks to the perceiving efficiency.
With only 0.8M total video samples, our model outperforms contemporary LMMs with the same backbone and comparable parameters amount.
\textbf{An experiment in A/B pretraining with same backbone and training data shows that, w/ cubing achieves lower training loss and higher benchmark accuracy (+2.1) than the fixed partitioning baseline, strictly validating the benefits.}
Moreover, the front-loaded compression in encoding substantially reduces latency in the computationally expensive prefill and decoding, making it suited to mainstream EPD disaggregated inference deployment.
\url{https://quicksviewer.github.io}
\end{abstract}

\section{Introduction}
\label{sec:intro}

\begin{figure}[ht]
    \centering
    \renewcommand{\thesubfigure}{\arabic{subfigure}}
    \begin{subfigure}[b]{0.38\linewidth}
        \centering
        \includegraphics[width=\linewidth]{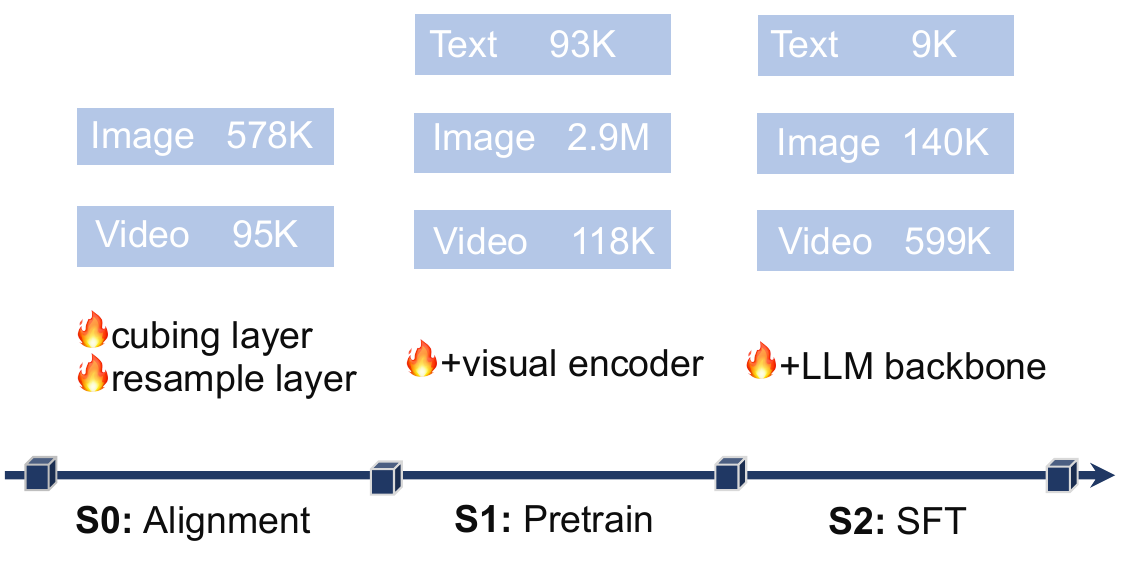}
        \label{fig:training_stages}
    \end{subfigure}\hfill
    \begin{subfigure}[b]{0.28\linewidth}
        \centering
        \includegraphics[width=\linewidth]{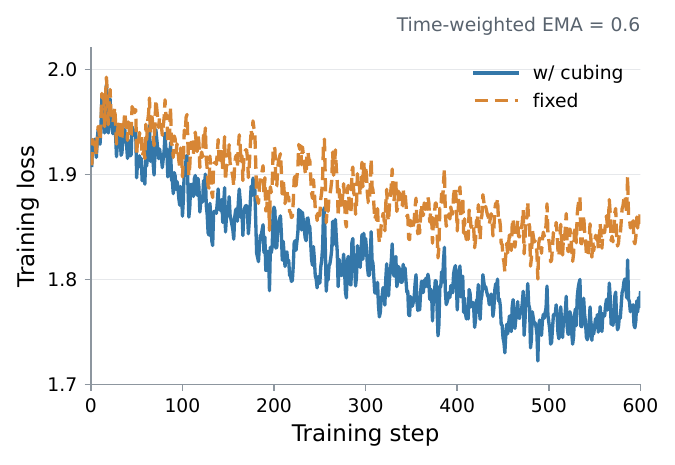}
        \label{fig:training_loss_comparison}
    \end{subfigure}\hfill
    \begin{subfigure}[b]{0.32\linewidth}
        \centering
        \includegraphics[width=\linewidth]{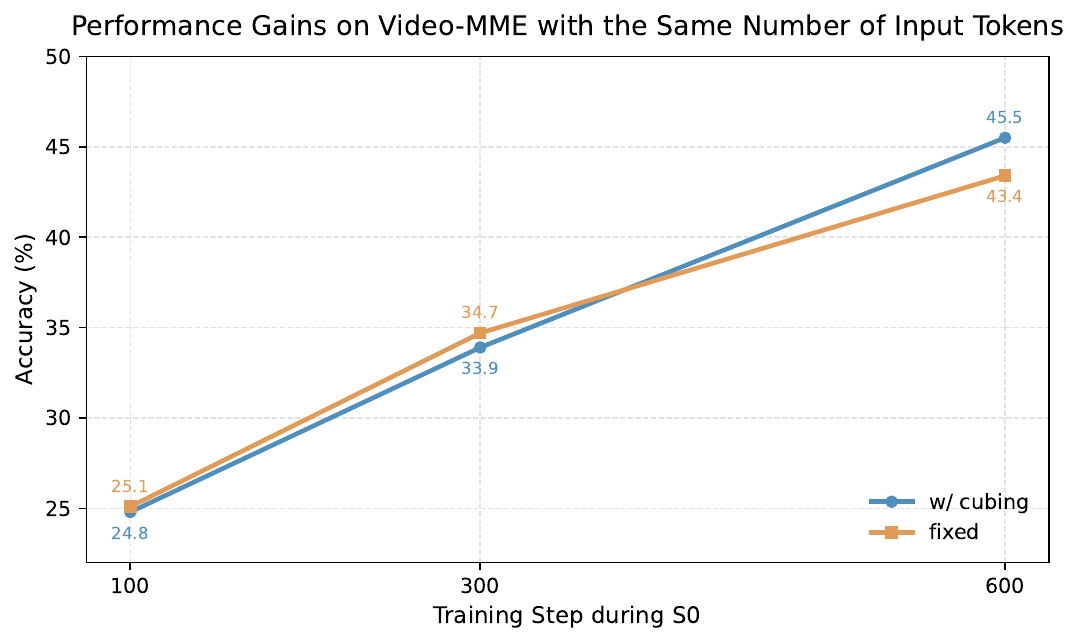}
        \label{fig:videomme_training_curve}
    \end{subfigure}
    \caption{\textbf{Left:} Three-stage training. \textbf{Middle and right:} Controlled A/B pretraining with the same backbone (Qwen2.5 + SigLIP) and training data (150K captions). Cubing yields lower training loss and improves benchmark accuracy (+2.1) over fixed partitioning at the same input token budget.}
    \label{fig:intro_fig}
\end{figure}

Video understanding with Large Multimodal Models (LMMs)~\citep{GoogleDeepmind2024Gemini2,OpenAI2024gpt4o,bai2025qwen2} is fundamental to intelligence in the physical world and requires efficient video perception that addresses the tension between the \textbf{temporal redundancy} of frames~\citep{buckler2018eva2,wenger1997video} and the \textbf{computational efficiency} of long-context models~\citep{fu2024challenges}.
Extensive studies have been striving to develop LMMs for solving this fundamental issue. Building on devise of frame sampling methods, most efforts typically involve dedicated token merging strategies~\citep{bai2025qwen2,wang2025internvideo2,shen2024longvu,zohar2024apollo,li2024videochat,zhang2025videollama} and adapted parallel training infra~\citep{zhang2024long,chen2024longvila,shen2025long}.
However, the arbitrary frames sampling and tokens merging introduces inevitable information loss, while marginal compression limits the number of frames in large-scale pre-training.

\textbf{Redundancy in video frames is inherently nonuniform over time, suggesting that the long, high-redundancy intervals and short, low-redundancy intervals should be encoded with an identical token budget.}
For example in Figure~\ref{fig:teaser_fig}, the first segment (frames 0-1) corresponds to a scene with rapid content changes, followed by a long period of stable visual content frames 2-6), and an almost static segment in the last part (frames 7-9).
However, it is nontrivial to determine intervals (\textit{i.e.,} cubes) to guarantee consistent information content and intact video events, so as to ultimately enhance LMM understanding.

We present Quicksviewer, an LMM that perceives nonuniform video cubes for efficient video understanding.
Given a video passed from a visual encoder, a small cubing network first partitions it into nonuniform cubes based on the \textbf{momentum} of semantic feature differences between frames, a process that can be conducted online in streaming scenarios. Next, a unified resampling is employed to the cubes to gather a fixed number of tokens for adaptive compression. Finally, these visual tokens, along with absolute timestamps, are fed into the LLM for inference.
We integrates the learning of the cubing network into the end-to-end training of the LMM using the Gumbel Softmax~\citep{jang2016categorical,herrmann2020channel} method with an improved noise annealing mechanism. This approach not only enables efficient learning on videos without boundary labels but also insures sufficient sampling over the cubes distribution with \textbf{continuous gradient} during stabilized training.
Further, we introduce a boundary-weighted thumbnail as a surrogate for discrete boundary optimization to enable continuous gradient flow; appended after the cube tokens, it also serves as a compact memory analogous to Mamba's recurrent state~\citep{gu2023mamba}, supporting long-video understanding and online accumulation during inference.

We train our models starting from the LLM backbones through three progressive stages, each incorporating lengthy videos averaging 420s/1fps by benefiting from the efficient perception mechanism. The resulting model, which we coined as Quicksviewer, is an efficient LMM capable of understanding single/multi-images and long videos.
With only 0.8M video-text samples in total for training, our model achieves outperformed performance on Video-MME~\citep{fu2024video} using just up to 5\% of the tokens per frame required by baselines.
We also conduct extensive ablation studies to show the effectiveness of presented components, including the cubing approach, 3D positional encoding, loss penalty, and the annealing strategy.
Moreover, as shown in~\ref{fig:teaser_fig}, the front-loaded compression in encoding stage substantially reduces latency in the computationally expensive prefill and decoding stages, making it well suited to current mainstream EPD disaggregated inference deployment.

\section{Related Works}
\label{sec:related}

\paragraph{LMMs for General Vision Understanding}
Early LMMs align pretrained vision encoders with language models through image-text training and visual instruction tuning~\citep{alayrac2022flamingo,li2023blip,liu2023visual}.
Further developments improve visual grounding and multimodal reasoning~\citep{bai2023qwenvl,wang2024cogvlm,qi2024cogcom,chen2024internvl}, enabling more detailed perception and complex visual question answering.
Recent extensions learn from video-text data to support video understanding~\citep{hong2024cogvlm2,chen2024expanding,wang2024qwen2,bai2025qwen2,li2024llava,yao2024minicpm,li2023videochat,li2024mvbench}.
These models commonly treat sampled video frames as images, encode them with a shared visual backbone, and aggregate their visual tokens for language-based reasoning.

\paragraph{LMMs for Efficient Video Understanding}
Efficient video LMMs reduce the cost of representing sampled frames at several stages.
Before LLM processing, Video-ChatGPT and PLLaVA use pooling to produce compact visual representations~\citep{maaz2023video,xu2024pllava}.
VideoLLaMA2, Kangaroo, Apollo, and BIMBA combine spatial and temporal compression through convolution, clip-level resampling, or learned aggregation~\citep{cheng2024videollama,liu2024kangaroo,zohar2024apollo,islam2025bimba}.
Adaptive methods such as LongVU and TimeMarker allocate visual tokens according to video length, content, or instructions~\citep{shen2024longvu,survey_2411_18211}.

Other methods compress information within LLM or retain a bounded visual memory.
VideoStreaming and Video-XL illustrate memory aggregation or compressed-context approaches~\citep{survey_2404_05726,survey_2405_16009,shu2025video,survey_2406_12275}.
Inference-time methods such as DyCoke, PruneVid, FrameFusion, and LLaVA-Scissor reduce tokens or cached states using temporal redundancy and token importance~\citep{survey_2411_15024,survey_2412_16117,survey_2501_01986,survey_2505_21334,survey_2506_21862}.
Appendix~\ref{app:detailed_literature_review} provides the detailed literature review.

\begingroup
\hypersetup{pdfborder={0 0 0}}
\captionsetup{position=bottom,skip=6pt}
\scriptsize
\setlength{\tabcolsep}{1.5pt}
\setlength{\LTcapwidth}{\textwidth}
\renewcommand{\arraystretch}{1.1}
\newcolumntype{P}[1]{>{\raggedright\arraybackslash}p{#1}}
\newcolumntype{C}[1]{>{\centering\arraybackslash}p{#1}}
\begin{longtable}{@{}P{3.35cm}C{1.9cm}C{1.75cm}C{1.25cm}C{0.9cm}C{1.25cm}C{1.7cm}@{}}
\toprule
Model & Visual tokens / original frame & Compression & Max train frames & Eval\newline frames & $D_{600}$ (\%) & Video-MME\newline w/o subs (\%) \\
\midrule
\endfirsthead
\toprule
Model & Visual tokens / original frame & Compression & Max train frames & Eval\newline frames & $D_{600}$ (\%) & Video-MME\newline w/o subs (\%) \\
\midrule
\endhead
\bottomrule
\endfoot
\bottomrule
\caption{Comparison of Quicksviewer with representative video LMMs and compression methods. Metrics are compiled from official implementations or reported results; see Appendix~\ref{app:detailed_literature_review} for definitions and details. Quicksviewer achieves 45$\times$ compression and 62.3\% accuracy on Video-MME.}
\label{tab:video_compression_comparison}\\
\endlastfoot
\multicolumn{7}{l}{\textbf{Trained compression models}}\\*
\hyperlink{cite.cheng2024videollama}{\textcolor{blue}{VideoLLaMA2 7B}}\nocite{cheng2024videollama} & 72 & 8x & 16 & 16 & 97.33 & 47.9 \\
\hyperlink{cite.xu2024pllava}{\textcolor{blue}{PLLaVA 7B}}\nocite{xu2024pllava} & 144 & 4x & 16 & 16 & 97.33 & - \\
\hyperlink{cite.liu2024kangaroo}{\textcolor{blue}{Kangaroo 8B}}\nocite{liu2024kangaroo} & 128 & 8x & 160 & 64 & 89.33 & 56.0 \\
\hyperlink{cite.survey_2411_18211}{\textcolor{blue}{TimeMarker 8B}}\nocite{survey_2411_18211} & 36* & 16x* & 128 & 120* & 80.00* & 57.3 \\
\hyperlink{cite.shen2024longvu}{\textcolor{blue}{LongVU 7B}}\nocite{shen2024longvu} & $\leq$144 & Adaptive & - & 1 FP & - & - \\
\hyperlink{cite.shu2025video}{\textcolor{blue}{Video-XL 7B}}\nocite{shu2025video} & 144 & 4x pooling & 128 & 128 & 78.67 & 55.5 \\
\hyperlink{cite.islam2025bimba}{\textcolor{blue}{BIMBA 7B}}\nocite{islam2025bimba} & 144 & 4x & 128 & 128 & 78.67 & - \\
\hyperlink{cite.zohar2024apollo}{\textcolor{blue}{Apollo 7B}}\nocite{zohar2024apollo} & 32 & 22.78x & 600 & - & - & 61.3 \\
\multicolumn{7}{l}{\textbf{Reference video models}}\\*
\hyperlink{cite.zhang2024long}{\textcolor{blue}{LongVA 7B}}\nocite{zhang2024long} & 144 & 4x & N/A & 128 & 78.67 & 52.6 \\
\hyperlink{cite.chen2024longvila}{\textcolor{blue}{LongVILA 7B-256f}}\nocite{chen2024longvila} & 256 & 4x & 256 & 256 & 57.33 & 60.1 \\
\hyperlink{cite.survey_minicpm_v26}{\textcolor{blue}{MiniCPM-V 2.6}}\nocite{survey_minicpm_v26} & 64 & 16x square* & - & - & - & 60.9 \\
\hyperlink{cite.li2024llava}{\textcolor{blue}{LLaVA-OneVision 7B}}\nocite{li2024llava} & 196 & 3.72x & 32 & 32 & 94.67 & 58.2 \\
\multicolumn{7}{l}{\textbf{Inference-time methods}}\\*
\hyperlink{cite.survey_2411_15024}{\textcolor{blue}{DyCoke 7B}}\nocite{survey_2411_15024} & 122.5 & 1.6x pooled & N/A & 32 & 94.67 & 59.5 \\
\hyperlink{cite.survey_2411_14401}{\textcolor{blue}{DyTo 7B}}\nocite{survey_2411_14401} & 36.8 & 15.65x & N/A & 100 & 83.33 & 41.2 \\
\hyperlink{cite.survey_2506_02850}{\textcolor{blue}{METoK 7B}}\nocite{survey_2506_02850} & Adaptive & Adaptive & N/A & 32 & 94.67 & 58.4 \\
\midrule
\textbf{Quicksviewer 8B (Ours)} & 12.8 & 45x & 420 & 420 & 30 & 62.3 \\
\end{longtable}
\endgroup

\section{Approach}
\label{sec:method}

\begin{figure*}[ht]
    \centering
    \includegraphics[width=\textwidth]{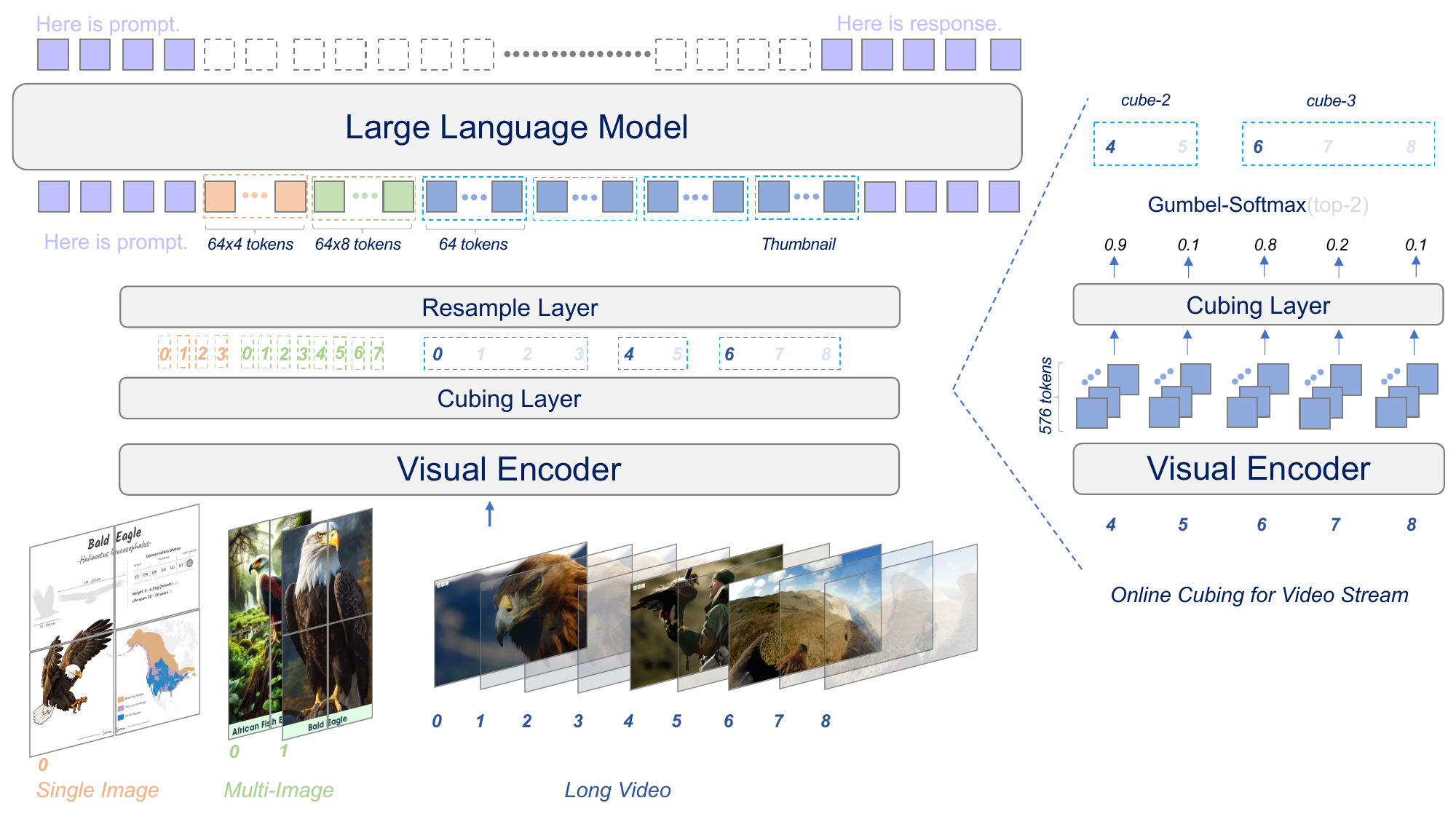}
    \caption{Left: The network architecture of Quicksviewer, that performs unified understanding of videos and images through visual tokens from cascaded modules. Right: The cubing network, that partitions an online video into nonuniform cubes based on Gumbel Softmax.}
  \label{fig:framework}
\end{figure*}

We build an efficient LMM that can receive both images and videos.
The model consists of four basic components: (1) a visual encoder $f_{v}(\cdot)$ that encode image slices or video frames into visual tokens, (2) a cubing network $f_{q}(\cdot)$ that partitions the video frames into \textbf{$N_Q$}
cubes, (3) a resampler $f_{r}(\cdot)$ which compress each slice or cube into a fixed number of tokens, and (4) an LLM $f_{l}(\cdot)$ which accept concatenation of visual tokens and user prompt for response generation.
Note that we introduce \textbf{FPQ}, the average number of frames per cube, which regulates the percepion granularity and enables adaptive number of cubes \textbf{$N_Q$} with duration.

\subsection{Cubing using Gumbel Softmax}
\label{subsec:cubing_using_gumbel_softmax}

\paragraph*{Visual Encoding}
Given an input video to our model, we first uniformly sample $N_F$ frames with a fixed FPS to form $[F_i]_{i=1,..,N_F}$. And then, each frame is firstly encoded using the visual encoder $f_v$ to obtain $N_1$ visual tokens $\mathbf{F}_j=[\mathbf{v}_i]_{i=1,..,N_1}$, where $\mathbf{v}_i\in\mathbb{R}^{d}$ is the token representation.
The large number of visual tokens provides fine-grained semantic representations for each frame, while the concatenation of all tokens (\emph{i.e.,} $N_F\times N_1$) across frames results in an unacceptable sequence length.

\vspace{-5pt}
\paragraph*{Cubing} It is natural to leverage the semantic difference of features to find keyframes.
For compatibility with online streaming scenarios and also considering the event-level changes, we utilize the momentum accumulated from the previous frames to track the semantic changes. Given the visual tokens of frames, the cubing network calculates the $i^{th}$ momentum representation:

\vspace*{-12pt}
\begin{align}
\Delta_i = \alpha(\mathbf{F}_i - \mathbf{F}_{i-1}) + (1-\alpha)\Delta_{i-1}
\end{align}

where $\Delta_i=[\mathbf{\delta}_j]_{j=1,..,N_1}\in\mathbb{R}^{N_1\times d}$ captures the accumulated semantic changes discounted by factor $\alpha\in[0,1]$.
A two-layer LayerNorm MLP~\citep{ba2016layer} then maps the mean visual-token momentum to a frame significance score:

\vspace{-20pt}
\begin{align}
\mathbf{y}_i = \textrm{MLP}_{\times 2}(\textrm{LayerNorm}(\frac{1}{N_1}\sum_{j=1}^{N_1}\delta_j)), \qquad \mathbf{y}_i\in\mathbb{R}^{2}
\end{align}
\vspace{-18pt}

where $\mathbf{y}_i=[w_{i,0},w_{i,1}]$ contains logits for non-boundary (class 0) and boundary (class 1). Without noise and at $\tau=1$, the sigmoid gives the keyframe probability $p_i=\sigma(w_{i,1}-w_{i,0})$~\citep{herrmann2020channel}.
We apply \textit{top-k} to the post-softmax class-1 probabilities $z_{i,1}$ from Eq.~(\ref{eq:gumbel_sampling}) along the temporal dimension ($i=2,\ldots,N_F$), selecting $N_Q-1$ boundary indices $[l_i]_{i=1,..,N_Q-1}$ in temporal order. The keyframe $F_{l_i}$, along with its subsequent consecutive non-keyframes, forms the cube $Q_{i+1}=[F_{l_i}, .., F_{l_{i+1}})$.
Note that the first sampled frame consistently serves as the keyframe for deriving the first cube $Q_1$.

\paragraph*{Sampling for Training}

During training, we use Gumbel-Softmax~\citep{gumbel1954statistical,jang2016categorical} to explore boundary selections. The relaxed probabilities are:

\vspace{-20pt}
\begin{align}
\mathbf{z}_i = \operatorname{softmax}\!\left(\frac{\mathbf{y}_i+\eta_t\mathbf{g}_i}{\tau}\right), \qquad g_{i,c}\overset{\mathrm{i.i.d.}}{\sim}\mathrm{Gumbel}(0,1)
\label{eq:gumbel_sampling}
\end{align}

where $c\in\{0,1\}$, $\tau>0$ controls softmax sharpness, and $\eta_t$ scales the noise ($\eta_t=0$ at inference). The resulting binary boundary indicators $b_i$ use $z_{i,1}$ as the straight-through gradient surrogate for $i\ge2$; $b_1=1$.

In experiments, we found that persistent exploration prevents the model from establishing a stable cubing paradigm for subsequent reasoning, leading to sustained loss oscillations.
We therefore \textbf{anneal} the noise scale $\eta_t$ from $\eta_{0}=1.0$ to $\eta_{T}=0.001$ using a cosine schedule.

\subsection{Resampling with 3D Positional Encoding}

Based on the partitions from cubing network, a unified 3D resampler is adopted to compress each cube of arbitrary length into a fixed number of $N_2$ tokens.

\paragraph*{Resampling Video Cubes}

Following~\citet{yao2024minicpm}, we use a Perceiver resampler with one multi-head cross-attention layer. For cube $Q_c$, we flatten all frame-patch features into a sequence $\mathbf{X}_c$ and use $N_2=64$ learned queries $\mathbf{U}$ to aggregate them:
\[
\mathbf{H}_c = \operatorname{MHA}(\mathbf{U},\mathbf{X}_c+\mathbf{P}_c,\mathbf{X}_c).
\]
The arguments denote queries, keys, and values, respectively; input projections and normalization are omitted for clarity.
We extend the original 2D positional encoding to $\mathbf{P}_c$ with 3D coordinates $(x,y,z)$ denoting time, width, and height.
The attention outputs are normalized and projected into the LLM embedding space, yielding 64 tokens irrespective of cube duration.
For images, we first adopt the AnyRes~\citep{liu2024llavanext} to divide high resolutional images into slices, and apply the same resampling to each slice to obtain visual tokens and finally concatenate the tokens.

\paragraph*{Resampling for Video Thumbnail}

Using cubes quantized from the cubing network for response tokens generation introduces a fundamental problem: \textbf{\itshape How does the NTP training objective optimize the boundary prediction of the cubing network?}
We introduce video thumbnail to resolve this problem meanwhile provide effective global representation.
Specifically, we weight the selected frame representations $\mathbf{F}_i$ by their straight-through boundary indicators $b_i$ and average them to obtain $N_1$ visual tokens.
A further resampling is performed to get final thumbnail representation containing $N_2$ tokens.
This simple approach allows gradients to be directly propagated back to the cubing boundaries.
Moreover, appended after the cube tokens, the thumbnail serves as a compact memory analogous to Mamba's recurrent state~\citep{gu2023mamba}, supporting long-video understanding and online accumulation of visual information during inference.

\subsection{LLM \& Optimization}

Given the visual tokens coming from videos or images, and the prompting tokens encoded from the LLM, the final sequence feed into the LLM is the concatenation of these tokens, separated by special indicator tokens and absolute timestamps as a float number in 0.01-second units for videos.

During training, we observed that excessively large logits $\mathbf{y}$ cause overly high gradients, impairing convergence. To address this, we introduce an auxiliary $L_2$ norm loss with $\beta=0.001$ penalty weight on them to constrain its values within a reasonable range.

\section{Training Process}
\label{sec:training_process}
\vspace{-10pt}

We perform three-stage training from LLM backbones, with videos averaging 420s in each stage.

\paragraph{Stage-1: Multimodal Alignment}
\label{parag:stage2-algin}

We utilize both interleaved and captioning image-text corpuses, and video-text captioning corpus to train our models, establishing fundamental alignment between visual encoder and LLM backbones with in-context learning capabilities.
We sample a subset of 20K sequences from OBELICS~\citep{laurenccon2023obelics}, with each containing more than two interleaved pairs.
We utilize LCS~\citep{li2024llava}, a re-captioned dataset consisting of 558K detailed descriptions from the CC3M~\citep{sharma2018conceptual}.
The video-text training data incorporates a sampled subset of 87K captioning pairs from FineVideo~\citep{Farre2024FineVideo} and 8K captaining pairs from ANetCaptions~\citep{krishna2017dense}.
We train parameters of the cubing network and resampler while keeping all other parameters frozen to establish a stable projection.
The models are trained for 1 epoch with a $lr$ that warms up to $1e^{-4}$ over the first 2\% of steps, then gradually decays to 0.

\paragraph{Stage-2: Pre-training}
\label{parag:stage2-pt}

We employ large-scale pretraining data, primarily consisting of image-text multi-task data, to pre-train models establishing general multimodal capabilities across broad visual scenarios.
We utilize a subset of 2.99M samples from LLaVA-OneVision-SingleImage~\citep{li2024llava} as training corpus, which incorporates 2.9M image-text pairs and 93K textual instruction-tuning samples from Evo-Instruct~\citep{chen2024allava}.
For video-text corpus, we utilize a sampled subset of 75K video QAs from FineVideo~\citep{Farre2024FineVideo} and 38K captioning pairs from ShareGPT4Video~\citep{chen2024sharegpt4video}.
To mitigate catastrophic forgetting, we retain 5\% of the previous image and video data in our training corpus.
Alongside the cubing network and resampler, we also unfreeze the visual encoder to improve the visual representation.
We train models for 1 epoch with a $1e^{-5}$ initial learning rate, with the same warmup and decay schedule as stage-1.

\paragraph{Stage-3: Supervised Fine-tuning}
\label{parag:stage3-sft}

We primarily utilize a subset of 476K video-text samples sourced from VideoChat2-IT~\citep{li2024mvbench}, and a subset of 79K samples from ShareGPTVideo~\citep{zhang2024direct} as the video corpus.
To enhance adaptation to long video scenarios, we further integrate 5K samples from MovieChat~\citep{song2024moviechat} and 39K samples derived by \citep{chen2024longvila} from the Shot2Story dataset~\citep{han2023shot2story20k}.
The image-text corpus incorporates a sampled subset of 100K mult-image, multi-task understanding samples from LLaVA-OneVision-MultiImages~\citep{li2024llava}.
We also preserve a subset of training data from the previous stage, consisting of 40K text-image pairs and 9K textual instruction-tuning samples.
We train all parameters for 1 epoch with a lr that warms up to $1e^{-5}$ over the 0.02 epoch, followed by a cosine decay to 0.

During training, all videos sampled at 1FPS to extract full frames. For videos exceeding 420s, we uniformly extract 420 frames.
The Gumbel noise learning rate $\eta$ (initialized at 1.0) undergoes cosine annealing to 0.001 within: 0.8 epoch (Stage 1) or 0.6 epochs (Stages 2-3).

\begin{table*}[ht]
    \centering
\begin{adjustbox}{width=\linewidth,center}
\renewcommand{\arraystretch}{1.2}
\setlength{\tabcolsep}{1.5mm}
\begin{tabular}{lcccccccccc}
\toprule \textbf{Models} & \textbf{Size}  & \textbf{\#Tokens} & \textbf{\#Train}  &  \textbf{MMBench-Video} & \textbf{MVBench} & \textbf{MLVU}  & \textbf{Video-MME} \\ \cline{7-8}
\rowcolor{gray!10} Duration & & /Frame & V-T  & 3 min &16 sec & 3$\sim$120 min & 1$\sim$60 min \\
\midrule
\textit{Proprietary Models} \\
GPT4-V~\citep{OpenAI2024gpt4v} & - & - & - & 1.53 & 43.7  & - & 60.7  \\
GPT4-o~\citep{OpenAI2024gpt4o} & - & - & - & 1.63 & 64.6  & 66.2 & 77.2 \\
\midrule
\textit{Open-Source Video LMMs} \\
LLaMA-VID~\citep{li2024llama} & 7B & 2 & 0.4M & 1.08 & 41.5  & 33.2 & - \\
LongLLaVA~\citep{wang2024longllava}  & 9B & 144 & 0.5M & - & 49.1 & - & 43.7 \\
Chat-UniVi~\citep{jin2024chat} & 7B  & 112 & 100K & 1.06 & 42.9 & - & 45.9 \\
ShareGPT4Video~\citep{chen2024sharegpt4video} & 8B & 144 & 4.8M & 1.05 & 51.2  & 46.4 & 43.6 \\
LLaVA-NeXT-Video~\citep{zhang2024llava} & 7B & 144 & 100K & 1.14 & 33.7 & - & 46.5 \\
VideoLLaMA2~\citep{cheng2024videollama} & 7B & 32 & 10.7M & 1.08 & 54.6  & 48.5 & 46.6 \\
LongVA~\cite{zhang2024long} & 7B & 144 & - & - & -  & 56.3 & 54.3 \\
VideoChat2~\citep{li2024mvbench} & 7B & 64 & 2.8M & 1.22 & 60.4  & 47.9 & 54.6 \\
mPLUG-Owl3~\citep{ye2024mplug} & 8B & 729 & 134K & \underline{1.35} & 54.5  &  - & 53.5  \\
Video-XL~\citep{shu2025video} & 7B & 9$\sim$72 & 0.1M & - & 55.3 & 45.5 & 55.5 \\
LongVILA~\citep{chen2024longvila} & 7B & 256 & 0.2M & - & 67.1 & - & 60.1 \\
LongVU~\citep{shen2024longvu} & 7B & 64 & 0.8M & - & 66.9 & 65.4 & \underline{60.6} \\
\midrule
\rowcolor{blue!10} \modelname & 8B & 12.8 & 0.8M & \textbf{1.37} &  62.3  & \textbf{66.1} & \textbf{62.3} \\
\bottomrule
\end{tabular}
\end{adjustbox}
\caption{Compared to baselines with similar model size, Quicksviewer achieves multiple SOTA performance while using fewer tokens per frame (up to 5\% of baseline) and less training samples.}
\label{tab:main}
\end{table*}

\section{Experiment}
\label{sec:experiment}

\subsection{Implementation Details}

\begin{figure*}[t]
    \centering
    \begin{subfigure}[t]{0.38\textwidth}
        \centering
        \includegraphics[width=\textwidth]{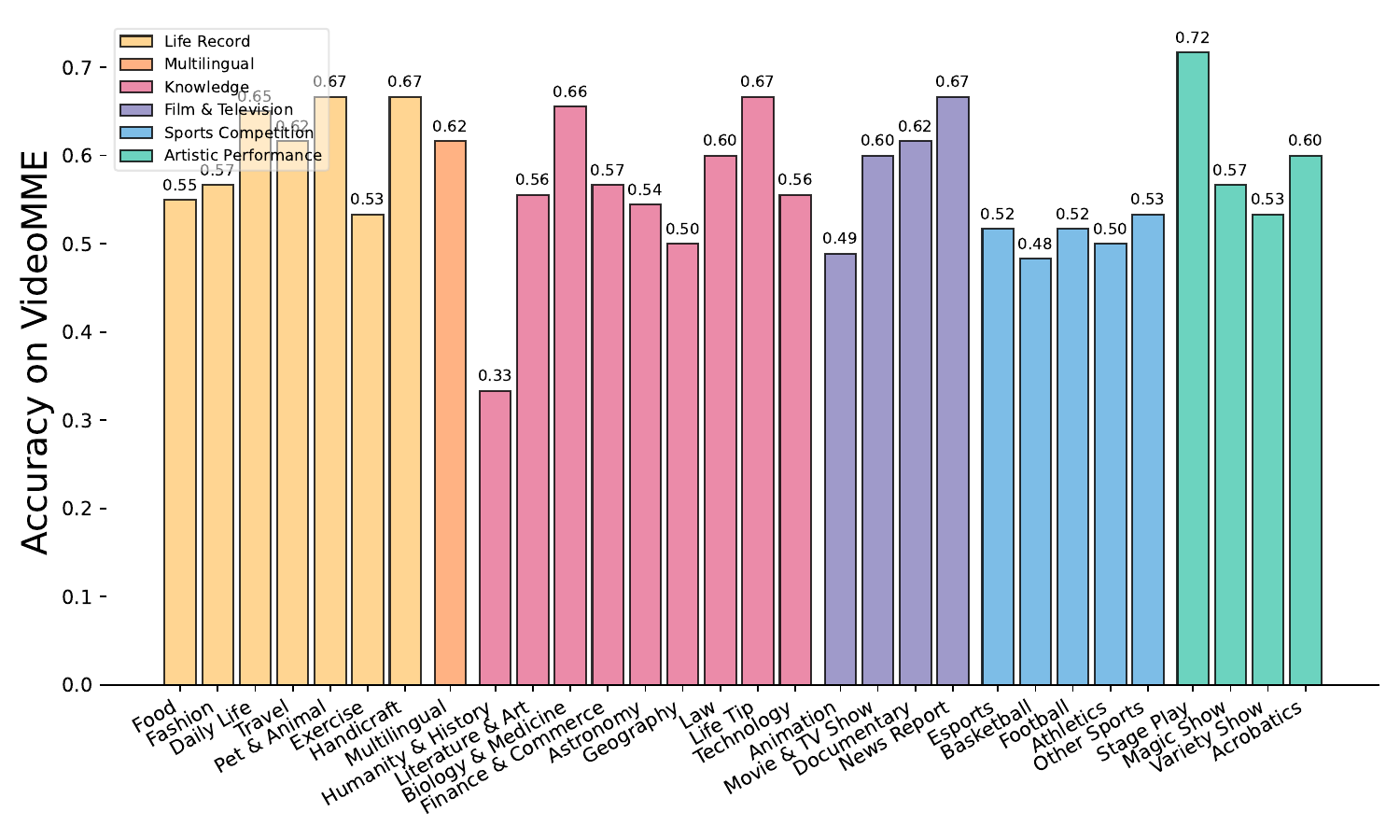}
    \end{subfigure}%
    ~
    \centering
    \begin{subfigure}[t]{0.22\textwidth}
        \centering
        \includegraphics[width=\textwidth]{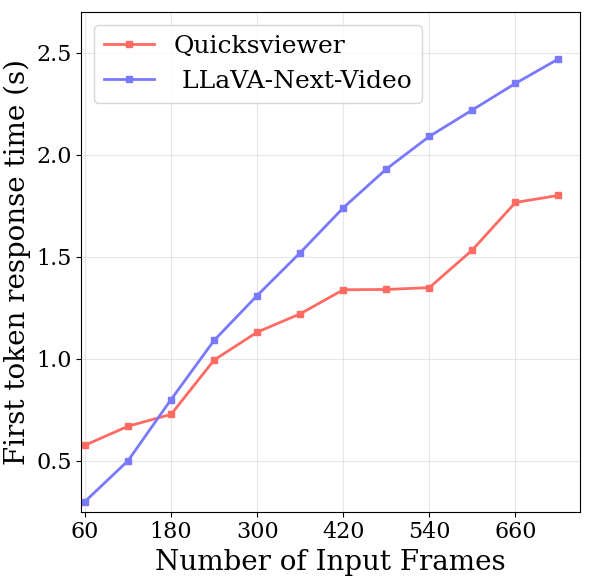}
    \end{subfigure}%
    ~ 
    \begin{subfigure}[t]{0.38\textwidth}
        \centering
        \includegraphics[width=\textwidth]{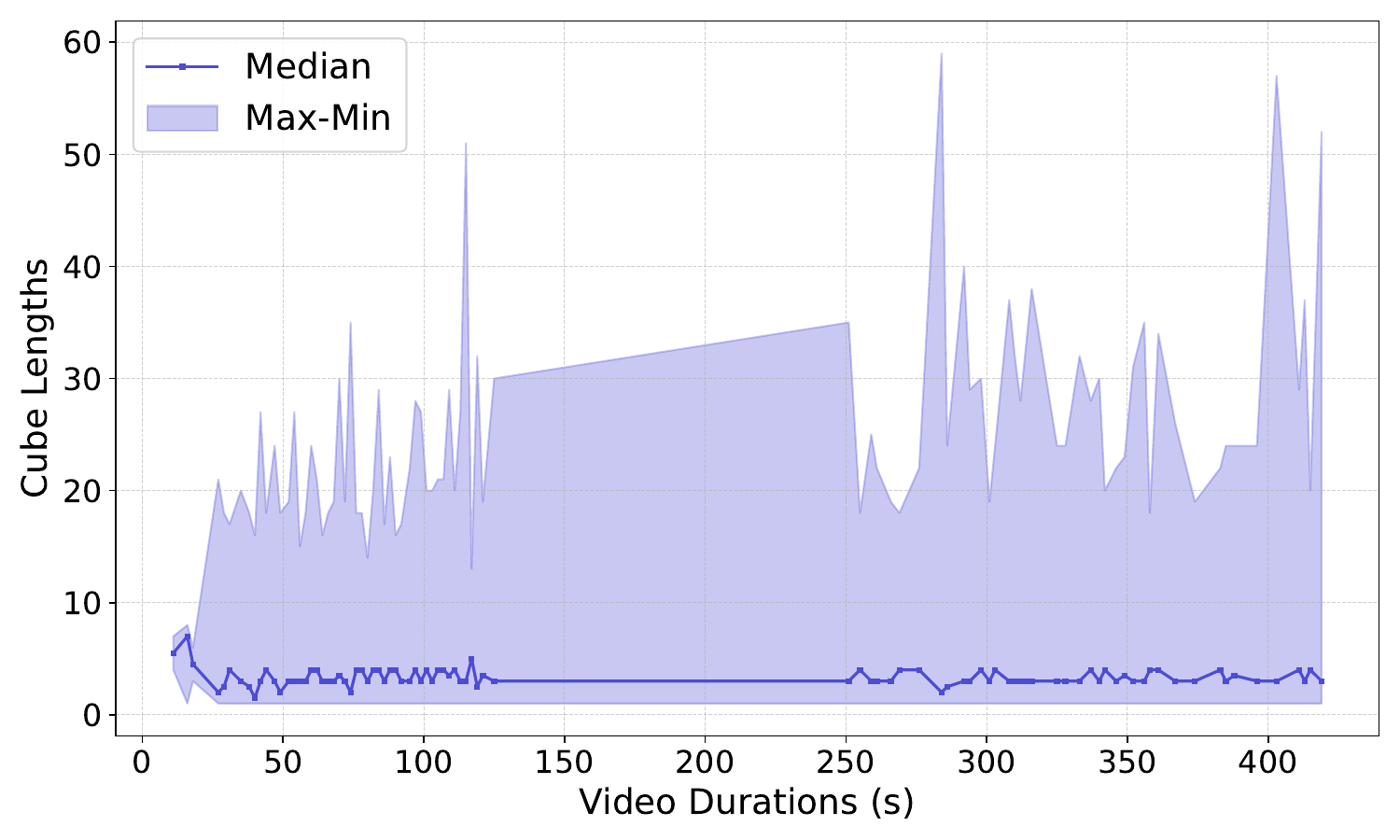}
    \end{subfigure}
    \caption{(a) Domain-specific results on VideoMME. (b) Inference time comparison with LLaVA-Next-Video. (c) Distribution of cube lengths on VideoMME.}
    \label{img:videomme_cats_cubelens}
\end{figure*}

\begin{figure*}[t]
    \centering
    \includegraphics[width=\textwidth]{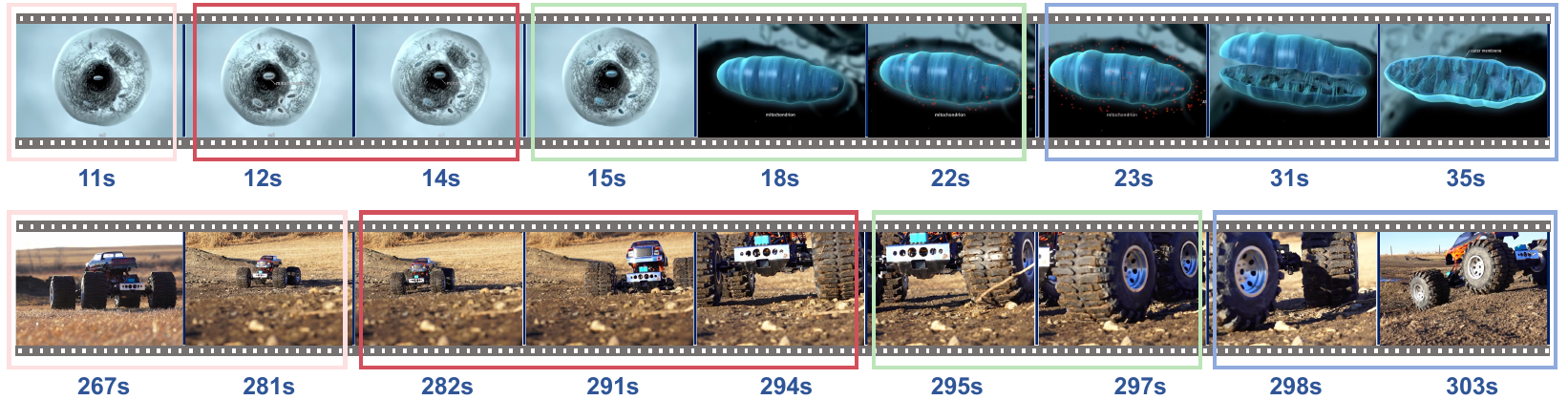}
    \caption{The \textbf{"Visual Lag"} phenomenon occurring during the model's cube-based segmental comprehension, where current cubes incorporate terminal frames from preceding event scenes to enable retrospective understanding.}
    \label{img:case_partition_cubes}
\end{figure*}

We use SigLIP~\citep{zhai2023sigmoid} (soo400m-path14-384) as our visual encoder in consistent with previous works.
We adopt Qwen2.5~\citep{yang2024qwen2-5} as the language backbone.
We use AdamW~\citep{loshchilov2017decoupled} optimizer with a cosine scheduler for all training stages.
The number of tokens generated from visual encoder and resampler are $N_1=576$, and $N_2=64$, respectively.
The discounting factor of momentum is set to $\alpha=0.9$.
The penalty weight to the auxiliary loss is set to $\beta=0.001$.
We use FPQ=5 for all stages.

\subsection{Experiments on Video Understanding}

We evaluate baselines with comparable sequence lengths and parameters for fair comparison.
For an unbiased comparison, we evaluate with baselines configured with comparable total sequence lengths, maintaining equivalent computational budgets.

\paragraph*{Benchmarks and Metrics}
We evaluate the our models on widely used video understanding benchmarks Video-MME~\citep{fu2024video}, MVBench~\citep{li2024mvbench}, and MLVU~\citep{zhou2024mlvu} to investigate the effectiveness.

\paragraph*{Quantitative Results}

We adopt empirically optimized configuration: 5 FPS with a maximum of 720 frames for all benchmarks.
As the main result shown in Table~\ref{tab:main}, our standard model achieves the best performance on Video-MME and MLVU, and the competitive performance on MVbench with significantly fewer tokens and training volumes.
Specifically, our model achieves optimal performance on Video-MME, albeit with slightly inferior results on long videos. This demonstrates that the encoding paradigm harness the scaling benefits by high frame rate.
Our approach achieves the best performance on MLVU while demonstrating competitive performance on MVBench, despite utilizing substantially less training data (only 28\% of VideoChat2's and 7.5\% of VideoLLaMA2's requirements).
This evidences our network's learning efficiency.

\begin{figure*}[ht]
    \centering
    \begin{subfigure}[t]{0.5\textwidth}
        \centering
        \includegraphics[height=1.4in]{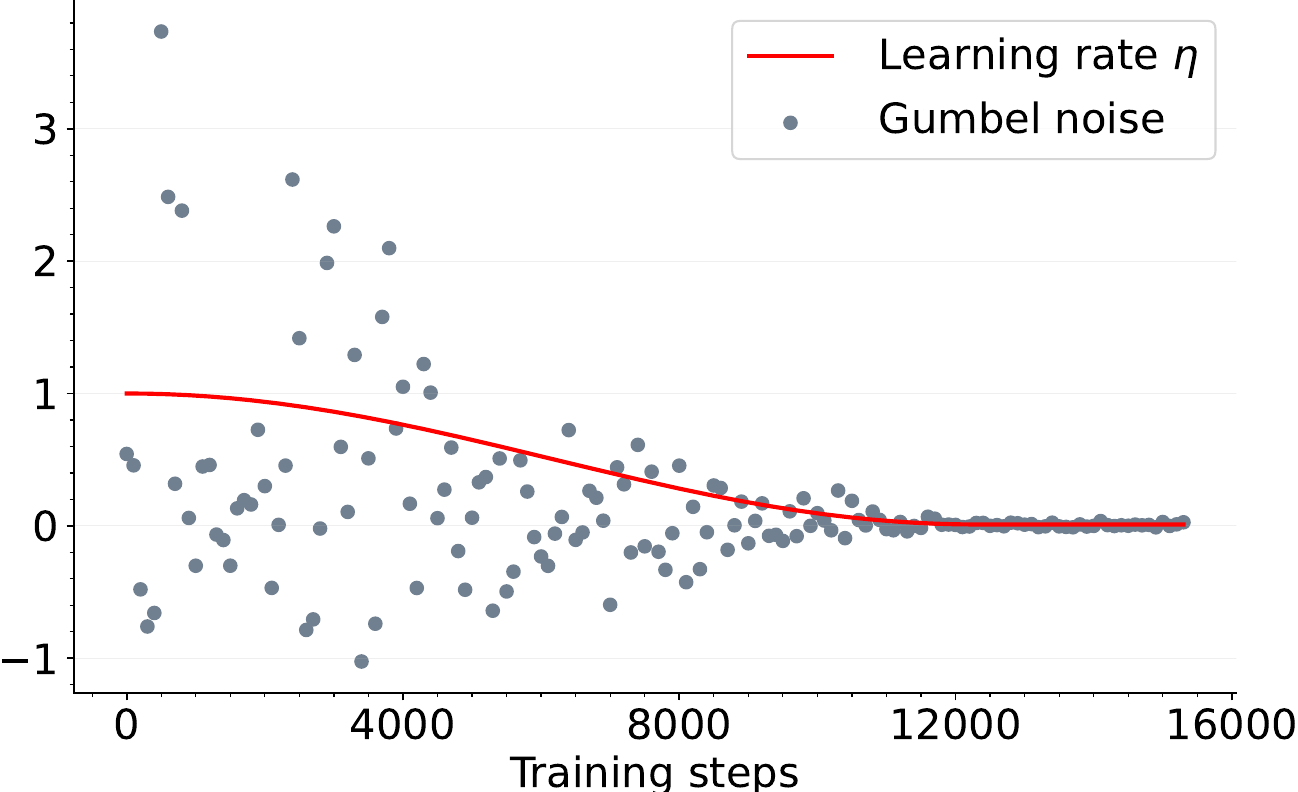}
    \end{subfigure}%
    ~ 
    \begin{subfigure}[t]{0.5\textwidth}
        \centering
        \includegraphics[height=1.4in]{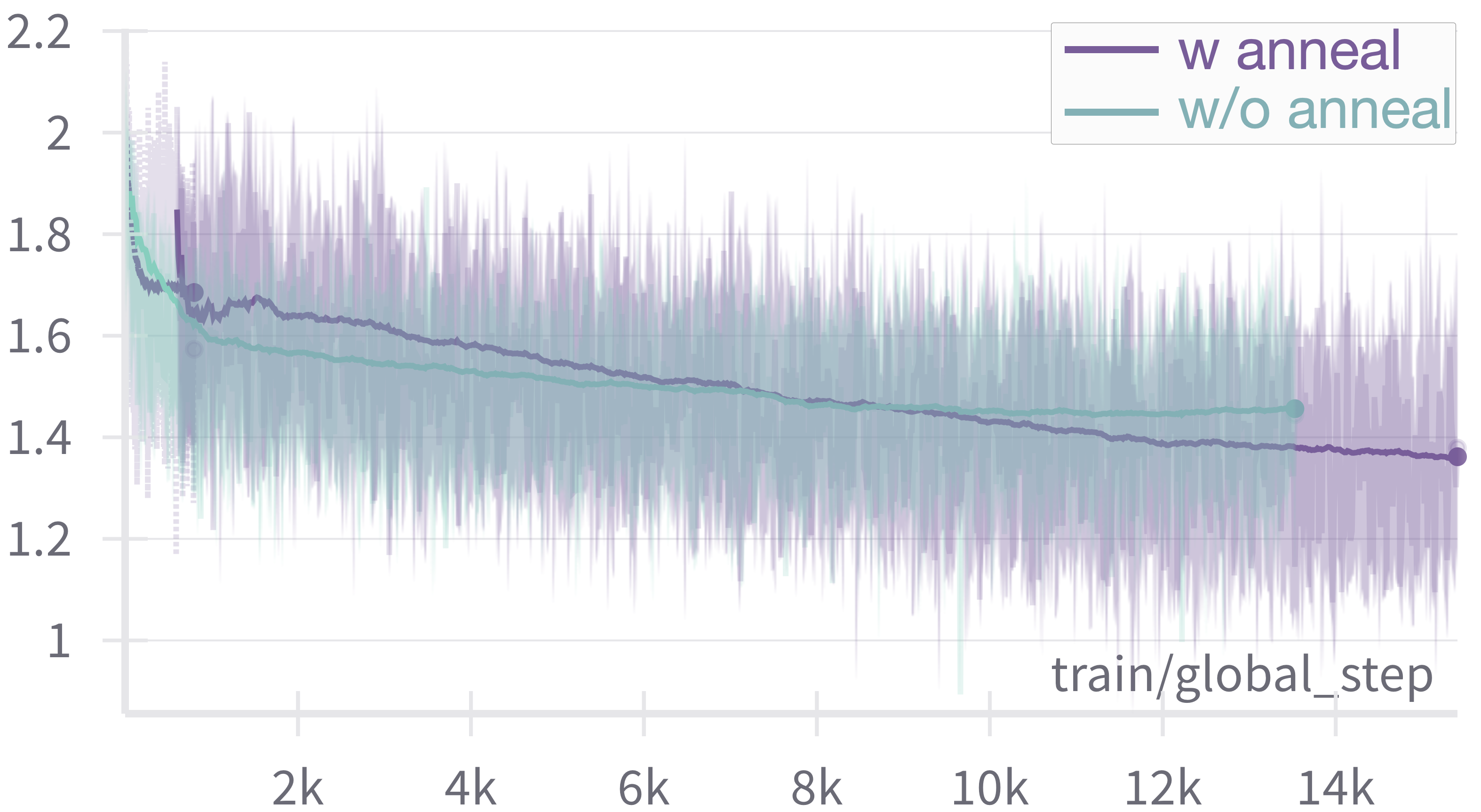}
    \end{subfigure}
    \caption{(a) Left: Gumbel noise progressively anneals to 0.001 following the decaying learning rate with cosine scheduler. (b) Right: Compared to non-annealed training (cyan curve), adding Gumbel noise annealing (purple curve) yields more stable and superior loss convergence.}
    \label{img:train_aneal_loss}
\end{figure*}

\paragraph*{Analysis}
We further analyze the model performance across distinct domain categories in Video-MME, systematically examine both capability advantages and limitations.
As illustrated in Figure~\ref{img:videomme_cats_cubelens} (a), bars sharing identical colors belong to the same domain.
Primarily, we observe consistent model performance across all domains, suggesting limited domain-specific variation in question difficulty.
Secondly, the model demonstrates suboptimal performance (below 50\%) in three categories: Humanity \& History, Animation, and Basketball. This may indicate persistent challenges in fine-grained character recognition that require further improvement.

We further analyze the distribution of cube lengths on Video-MME, with results shown in Figure~\ref{img:videomme_cats_cubelens} (b). Based on the predefined FPQ, we found the median cube length approximates 5 frames. Notably, the model demonstrates a tendency to partition diverse length of cubes for longer videos, which aligns with the variable viewing speeds in human perception of lengthy videos.

\subsection{Analysis of the Cubes Partitioning}

To investigate how the trained model  partitions cubes for understanding, we analyzed two representative video cases by examining cubes relative to content transitions. As shown in Figure~\ref{img:case_partition_cubes}, each box refers a cube spanning from its start to end timestamps (with similar intermediate frames omitted).
We reveal a \textbf{"Visual Lag"} phenomenon during cube-based video perception of the model: terminal frames from preceding event scenes are incorporated into cubes containing subsequent event scenes.
For example in the first video, the initial frames of cubes 2-4 respectively contain content from three event scenes: 1) cellular details, 2) mitochondrial positioning, and 3) ATP synthesis exhibiting in cubes 1-3 respectively.
We posit this mechanism enables the model to retain partial memory of preceding scenes to facilitate current scene understanding.

\subsection{Analysis of the Annealing Strategy}

\begin{table*}[t]
    \centering
\begin{adjustbox}{width=\linewidth,center}
\small
\renewcommand{\arraystretch}{1.2}
\setlength{\tabcolsep}{4mm}
\begin{tabular}{lcccccc}
\toprule  \multicolumn{5}{c}{\centering \textbf{Ablation Components}}  & \multicolumn{2}{c}{\centering \textbf{Video-MME} } \\ \cline{6-7}
 &&&& & Overall & Long \\
  \rowcolor{gray!10} \textbf{PE} & \textbf{Cubing Network} & \textbf{Penalty} & \textbf{Annealing} & \textbf{Trainable} &  1$\sim$60 min & 30$\sim$60 min \\
\midrule
  2D & ViT first 2-layers & 0.1 & - & $\theta_{c},\theta_{r},\theta_{v}$ & 33.92 & 35.56 \\
  2D & ViT full &  0.1  & -       & $\theta_{c},\theta_{r},\theta_{v}$ & 41.22 & 38.67 \\
  3D & ViT full &  0.1  & -       & $\theta_{c},\theta_{r},\theta_{v}$ & 44.37 & 40.67 \\
  3D & ViT full & 0.001 & -        & $\theta_{c},\theta_{r},\theta_{v}$ & 44.66 & 40.44 \\
  3D & ViT full & 0.001 & annealing & $\theta_{c},\theta_{r},\theta_{v}$ & 45.44 & 43.44 \\
 \rowcolor{blue!10}  3D & ViT full & 0.001 & annealing & All & \textbf{45.96} & \textbf{38.44} \\
\bottomrule
\end{tabular}
\end{adjustbox}
\vspace*{2pt}
\caption{Ablation results of Stage-3 training initialized from a checkpoint pretrained only with image data (Stage 1-2) for simplicity. The optimal configuration: 3D positional encoding, Gumbel noise annealing with 0.001 penalty weight, and full trainable parameters.}
\label{tab:ablation}
\end{table*}

Traditional Gumbel-Softmax training controls sampling randomness exclusively through temperature adjustment, making it unsuitable for training the cubing network as a component of an LMM. To resolve this issue, we propose annealing the Gumbel noise, which substantially improves both training stability and effectiveness.
To further evaluate the performance of the proposed annealing mechanism, we examine the evolution of Gumbel noise values throughout a training epoch in with the annealed learning rate. For clear visualization, we uniformly sample training steps at 100-step intervals, as illustrated in Figure~\ref{img:train_aneal_loss} (a).
From the figure, we observe that in the early training stages, larger Gumbel noise effectively facilitates exploration for the cubing network. As training progresses, the Gumbel noise gradually converges to the predefined value of 0.01. This allows the model to leverage its learned segmentation mechanism for video understanding in later stages, stabilizing the training process and achieving optimal performance.

Figure~\ref{img:train_aneal_loss} (b) compares the loss trajectories of models trained with and without the annealing mechanism. Initialized from the same checkpoint from Stage-2, we train parallel models using both approaches and monitor loss variations throughout one epoch to assess learning efficiency and stability.
Our analysis reveals that the model benefits from the progressive annealing of Gumbel noise in the later stages. During this phase, the model effectively utilizes its learned cubing mechanism to accelerate loss minimization, achieving superior convergence efficiency, demonstrating  the effectiveness and training stability.

\subsection{Ablation Studies}

We conduct comprehensive ablation studies to evaluate the efficacy of the components leveraged in \modelname. To establish a simple baseline, we first train a Llama3.1~\citep{touvron2023llama} model through Stages 1-2 using only image-text data introduce in Sec.~\ref{sec:training_process}, deliberately excluding video inputs. This image-only pretrained checkpoint then serves as the initialization point for systematically investigating various Stage-3 configurations with video-text data.

\paragraph*{Cubing network with ViT}

To accelerate cube processing, we investigate the feasibility of using only the initial n layers of ViT for the cubing network. Our ablation study employs the first 2 ViT layers for cube feature generation while maintaining all other model components unchanged. As demonstrated in Table~\ref{tab:ablation}, this configuration results in significant performance degradation, indicating that shallow visual features are insufficient for effective cubes partitioning.

\paragraph*{3D positional Encoding}

We systematically evaluate the impact of 3D positional encoding compared to the original 2D formulation. Implementing this modification while keeping all other parameters fixed in Stage-3 training, our experiments demonstrate a consistent accuracy improvement of +3.15\% (Table~\ref{tab:ablation}), confirming the benefits of spatiotemporal position awareness for video understanding.

\paragraph*{Penalty weight to the auxiliary loss}

The penalty weight applied to the $L_2$-norm of $\mathbf{y}$ values regulates the scale of logits, consequently influencing partition behavior. Through systematic ablation while maintaining other parameters fixed, we identify $\beta=0.001$ as the optimal weighting that simultaneously: (1) maintains $\mathbf{y}$ values in an effective operational range, and (2) achieves superior model performance (Table~\ref{tab:ablation}).

\paragraph*{The annealing strategy}

Having established the optimal penalty weight, we proceed to evaluate the efficacy of our proposed Gumbel noise annealing strategy. This approach systematically reduces exploration randomness during training, transitioning from aggressive parameter space exploration to fine-tuned optimization. Comparative results in Table~\ref{tab:ablation} demonstrate consistent performance improvements over the fixed-noise baseline, validating the benefits of noise scheduling.

\paragraph*{The tuning parameters}

We examine the impact of trainable parameters on video understanding by comparing two training regimes: (1) our baseline approach that only fine-tune the ViT and resampler parameters during Stage-3, versus (2) a full-parameter optimization strategy that additionally fine-tunes the LLM backbone. As evidenced in Table~\ref{tab:ablation}, comprehensive parameter training yields superior benchmark performance, suggesting that joint visual-linguistic optimization enhances multimodal alignment for video understanding tasks.

\section{Conclusion}
\label{sec:conclusion}

In this paper, we introduce Quicksviewer, an LMM designed for efficient video understanding through a nonuniform perception paradigm. By dynamically partitioning videos into nonuniform cubes and applying resampling, our approach achieves a 45$\times$ compression rate while maintaining a consistent representation for both images and videos. We demonstrated that integrating the cubing network into end-to-end training via Gumbel Softmax with an improved noise annealing mechanism, enables efficient learning without boundary labels.
Wen train model on 0.8M videos, attains superior performance on VideoMME with significantly fewer number of vision tokens.
Moreover, front-loading video compression into the encoding stage substantially reduces latency in the computationally expensive prefill and decoding stages, making Quicksviewer well suited to mainstream EPD disaggregated inference deployment.

\bibliography{iclr2027_conference}
\bibliographystyle{iclr2027_conference}

\newpage
\appendix
\section{Appendix}

\section{Training Data Details}
\label{app:train_data_details}

\subsection{Hyter-parameters}
\begin{table*}[h]
    \centering
    \setlength{\tabcolsep}{12pt}
    \renewcommand{\arraystretch}{1.2}
    \resizebox{\textwidth}{!}{%
    \begin{tabular}{@{}ll|c|c|c@{}}
    \toprule
    & & \textbf{Stage-1} & \textbf{Stage-2} & \textbf{Stage-3} \\
    \midrule 
    \multirow{2}{*}{\rotatebox[origin=c]{90}{\footnotesize \textit{Vision}}}
    & \textbf{Resolution}  & $384\times 384$ & $384\times 384$ & $384\times 384$ \\
    & FPS,\#Frames & 1, Max 420 & 1, Max 420 & 1, Max 420 \\
    \midrule 
    \multirow{4}{*}{\rotatebox[origin=c]{90}{\footnotesize \textit{Data}}}
    & \multirow{2}{*}{\begin{minipage}[l]{2cm}\textbf{Image-Text}\\\#samples\end{minipage}} & LCS, OBELICS & LLaVAOV-SingleImage & LLaVAOV-MultiImages  \\
    & & 558K, 20K & 2.99M & 100K \\
    & \multirow{2}{*}{\begin{minipage}[l]{2cm}\textbf{Video-Text}\\\#samples\end{minipage}} & FineVideo, ANetCaptions & FineVideo, ShareGPT4Video, ANetCaptions & Sec. 3.~\ref{parag:stage3-sft}  \\
    & & 87K, 8K & 118K & 599K \\
    \midrule
    \multirow{2}{*}{\rotatebox[origin=c]{90}{\footnotesize \textit{Model}}}
    & \textbf{Trainable} & Cubing, Resampler & Cubing, Resampler, ViT & Full Model  \\
    & \#Parameters & 75M & 500M & 8B \\
    \midrule 
    \multirow{3}{*}{\rotatebox[origin=c]{90}{\footnotesize \textit{Training}}}
    & \textbf{Anneal:} $\eta_{0},\eta_{T}$, ratio & 1.0, 0.01, 0.8 & 1.0, 0.01, 0.6 & 1.0, 0.01, 0.6 \\
    & \textbf{LR: $\theta_{c},\theta_{r},\theta_{v},\theta_{l}$} & $1e^{-4}$,$1e^{-4}$,-,- & $2e^{-5}$,$2e^{-5}$,$2e^{-5}$,- & $1e^{-5}$ \\    
    & \textbf{Epoch} & 1 & 1 & 1 \\
    \bottomrule
    \end{tabular}
    }
    \caption{Detailed configuration for each training stage.}
    \label{tbl:training_details}
    \end{table*}

\subsection{Training Data}

\begin{table}[ht]
\centering
\renewcommand{\arraystretch}{1.4}  
\setlength{\tabcolsep}{3mm}      
\small                      
\begin{tabular}{c|c|cc}
\toprule
\textbf{Modality} & \multicolumn{1}{c|}{\textbf{Task}} & \textbf{\# Samples} & \textbf{Dataset} \\ 
\hline
 \multirow{2}{*}{Image-Text} & Interleaved Pairs   & 20K & OBELICS \\
  & \cellcolor{gray!10} Single-Image Captioning   & \cellcolor{gray!10} 558K & \cellcolor{gray!10} LCS \\
\hline
 Video-Text  & Captioning     & 95K  & FineVideoCaptions, ANetCaptions \\
\bottomrule
\end{tabular}
\vspace{3mm}
\caption{Training data statistics for the alignment stage.}
\label{tab:app-traindata-alignment}
\end{table}

\begin{table}[ht]
\centering                         
\renewcommand{\arraystretch}{1.4}  
\setlength{\tabcolsep}{3mm}      
\small                      
\begin{tabular}{c|c|cc}
\toprule
\textbf{Modality} & \multicolumn{1}{c|}{\textbf{Task}} & \textbf{\# Samples} & \textbf{Dataset} \\ 
\hline
 Text &  Instruction     &  93K &  Evo-Instruct \\
\hline
\multirow{3}{*}{Image-Text} & \cellcolor{gray!10} Interleaved Pairs   & \cellcolor{gray!10} 20K & \cellcolor{gray!10} OBELICS \\
  &  Single-Image Captioning   &  50K & LCS \\
  & \cellcolor{gray!10} Single-Image Tasks   & \cellcolor{gray!10} 2.8M & \cellcolor{gray!10} LLaVAOneVision \\
\hline
 \multirow{3}{*}{Video-Text} & Captioning     & 5K  & FineVideoCaptions, AnetCaptions \\
 & \cellcolor{gray!10} VQA & 75K \cellcolor{gray!10}  & \cellcolor{gray!10} FineVideoQAs \\
 & Dense Captioning  &  38K & ShareGPT4Video \\
\bottomrule
\end{tabular}
\vspace{3mm}
\caption{Training data statistics for the pre-training stage.}
\label{tab:app-traindata-pretraining}
\end{table}

\clearpage
\begin{table}[ht]
\centering                         
\renewcommand{\arraystretch}{1.4}  
\setlength{\tabcolsep}{3mm}      
\small                    
\begin{tabular}{c|c|cc}
\toprule
\textbf{Modality} & \multicolumn{1}{c|}{\textbf{Task}} & \textbf{\# Samples} & \textbf{Dataset} \\ 
\hline
 Text & Instruction     & 9K &  Evo-Instruct \\
\hline
 \multirow{2}{*}{Image-Text} & \cellcolor{gray!10} Single-Image Tasks    & \cellcolor{gray!10} 40K & \cellcolor{gray!10} LLaVA-OneVision-SingleImage \\
 &  Multi-Images Tasks    & 100K &  LLaVA-OneVision-MultiImages \\
\hline
 \multirow{5}{*}{Video-Text} & Captioning     & 52K  & TextVR, MovieChat, YouCook2 \\
 & \cellcolor{gray!10} Dense Captioning & 4K \cellcolor{gray!10}  & \cellcolor{gray!10} ShareGPT4Video \\
 & Classification & 1K  & Kinetics-710 \\
 & \cellcolor{gray!10} VQA   & \cellcolor{gray!10} 354K  & \cellcolor{gray!10} \begin{tabular}[c]{@{}l@{}}\quad
 \quad NExT-QA, CLEVRER, EgoQA \\ TGIF, ShareGPTVideo, FineVideoQAs \end{tabular} \\
 & Instruction &  188K & VideoChatGPT, VideoChat, LongVILA \\ 
\bottomrule
\end{tabular}
\vspace{3mm}
\caption{Training data statistics for the supervised fine-tuning stage.}
\label{tab:app-traindata-sft}
\end{table}

\section{Detailed Literature Review}
\label{app:detailed_literature_review}

\paragraph{LMMs for Efficient Video Understanding}
Efficient video LMMs reduce the cost of representing sampled frames at several stages.
Before language-model processing, spatial pooling and learned resamplers produce compact visual representations, as in Video-ChatGPT, PLLaVA, LLaMA-VID, VideoLLaMA2, and LITA~\citep{maaz2023video,xu2024pllava,li2024llama,cheng2024videollama,survey_2403_19046}.
More recent approaches combine spatial and temporal compression through convolution, clip-level resampling, or learned aggregation, including Kangaroo, Apollo, PVC, Clapper, and BIMBA~\citep{liu2024kangaroo,zohar2024apollo,survey_2412_09613,survey_2505_15529,islam2025bimba}.
Adaptive methods allocate visual tokens according to video length, content, or instructions: representative examples include LongVU, TimeMarker, Dynamic-VLM, VideoLLaMA3, HICom, LinVT, and FiLA-Video~\citep{shen2024longvu,survey_2411_18211,survey_2412_09530,zhang2025videollama,survey_2503_16036,survey_2412_05185,survey_2504_20384}.

Other methods compress information within the language model or retain a bounded visual memory.
MA-LMM, VideoStreaming, Video-XL, and VoCo-LLaMA illustrate memory aggregation or compressed-context approaches~\citep{survey_2404_05726,survey_2405_16009,shu2025video,survey_2406_12275}.
Inference-time methods such as DyCoke, PruneVid, FrameFusion, ReTaKe, HoliTom, and LLaVA-Scissor reduce tokens or cached states using temporal redundancy and token importance~\citep{survey_2411_15024,survey_2412_16117,survey_2501_01986,survey_2412_20504,survey_2505_21334,survey_2506_21862}.
These methods differ in whether compression precedes the first language-model layer, occurs in deeper layers, or affects only the retained cache; the resulting token budgets therefore measure different costs.
We compare these mechanisms separately and report training frames, evaluation sampling, compression denominators, and subtitle protocols alongside benchmark accuracy.

\begingroup
\hypersetup{pdfborder={0 0 0}}
\captionsetup{position=bottom,skip=6pt}
\scriptsize
\setlength{\tabcolsep}{1.5pt}
\setlength{\LTcapwidth}{\textwidth}
\renewcommand{\arraystretch}{1.1}
\newcolumntype{P}[1]{>{\raggedright\arraybackslash}p{#1}}
\newcolumntype{C}[1]{>{\centering\arraybackslash}p{#1}}
\begin{longtable}{@{}P{3.35cm}C{1.9cm}C{1.75cm}C{1.25cm}C{1.55cm}C{1.7cm}C{1.7cm}@{}}
\toprule
Model & Visual tokens / original frame & Compression & Max train frames & Eval\newline frames & $D_{600}$ (\%) & Video-MME\newline w/o subs (\%) \\
\midrule
\endfirsthead
\toprule
Model & Visual tokens / original frame & Compression & Max train frames & Eval\newline frames & $D_{600}$ (\%) & Video-MME\newline w/o subs (\%) \\
\midrule
\endhead
\bottomrule
\endfoot
\bottomrule
\caption{Comparison of representative efficient video LMMs publicly available by June 30, 2025. Video-MME results use the no-subtitle setting. BIMBA follows the v1 128-frame setup (training cap assumed). Video-XL's count of 144 tokens/frame excludes additional visual summarization tokens (VSTs). \texttt{-} denotes unavailable or unresolved values, including unverified subtitle protocols; N/A denotes not applicable. $D_{600}=100[1-\min(600,F_{\mathrm{eval}})/600]$ is the sampling-budget omission for 600 input frames, not an architectural limit or an observed adaptive frame-drop rate. \texttt{Capacity0} means the reported frame cap is at least 600; ``+ adaptive'' indicates additional content-dependent pruning. Formatting tokens are excluded. Asterisks mark conditional settings: TimeSuite's ratio uses the raw spatial patch grid; TimeMarker's values use a 600-second video; MiniCPM-V's ratio assumes a single square 448-pixel view; ReTaKe's capacity does not account for adaptive frame selection.}
\label{tab:app_video_compression_comparison}\\
\endlastfoot
\multicolumn{7}{l}{\textbf{Trained compression models}}\\*
\hyperlink{cite.li2024llama}{\textcolor{blue}{LLaMA-VID 7B}}\nocite{li2024llama} & 2 & 128x & - & 1 FPS & conditional 0 & - \\
\hyperlink{cite.cheng2024videollama}{\textcolor{blue}{VideoLLaMA2 7B}}\nocite{cheng2024videollama} & 72 & 8x & 16 & 16 & 97.33 & 47.9 \\
\hyperlink{cite.zhang2025videollama}{\textcolor{blue}{VideoLLaMA3 7B}}\nocite{zhang2025videollama} & Dynamic & Dynamic & 180 & $\leq$180 & $\geq$70.00 & 66.2 \\
\hyperlink{cite.xu2024pllava}{\textcolor{blue}{PLLaVA 7B}}\nocite{xu2024pllava} & 144 & 4x & 16 & 16 & 97.33 & - \\
\hyperlink{cite.jin2024chat}{\textcolor{blue}{Chat-UniVi 7B}}\nocite{jin2024chat} & 7 at F=64 & 36.57x & 64 & 64 & 89.33 & - \\
\hyperlink{cite.survey_2403_19046}{\textcolor{blue}{LITA 7B}}\nocite{survey_2403_19046} & 3.56 at F=100 & 71.91x & 100 & 100 & 83.33 & - \\
\hyperlink{cite.survey_2404_03384}{\textcolor{blue}{LongVLM 7B}}\nocite{survey_2404_03384} & 3.05 at F=100 & 83.93x & 100 & 100 & 83.33 & - \\
\hyperlink{cite.survey_2410_19702}{\textcolor{blue}{TimeSuite / VideoChat-T}}\nocite{survey_2410_19702} & 3 & 65.33x* & 192 & 128 & 78.67 & 46.3 \\
\hyperlink{cite.liu2024kangaroo}{\textcolor{blue}{Kangaroo 8B}}\nocite{liu2024kangaroo} & 128 & 8x & 160 & 64 & 89.33 & 56.0 \\
\hyperlink{cite.survey_2411_18211}{\textcolor{blue}{TimeMarker 8B}}\nocite{survey_2411_18211} & 36* & 16x* & 128 & 120* & 80.00* & 57.3 \\
\hyperlink{cite.shen2024longvu}{\textcolor{blue}{LongVU 7B}}\nocite{shen2024longvu} & $\leq$144 & Adaptive & - & 1 FPS + pruning & - & - \\
\hyperlink{cite.shu2025video}{\textcolor{blue}{Video-XL 7B}}\nocite{shu2025video} & 144 & 4x pooling & 128 & 128 & 78.67 & 55.5 \\
\hyperlink{cite.survey_2501_03895}{\textcolor{blue}{LLaVA-Mini 8B}}\nocite{survey_2501_03895} & 1  +  fused text & Qualified & $<$60 reported & 1 FPS & conditional 0 & - \\
\hyperlink{cite.li2024videochat}{\textcolor{blue}{VideoChat-Flash 7B 448px}}\nocite{li2024videochat} & 16 & 49x input & 512 & $\leq$512 & $\geq$14.67 & 65.3 \\
\hyperlink{cite.survey_2503_16036}{\textcolor{blue}{HICom 7B}}\nocite{survey_2503_16036} & 20.5 at F=128 & 35.56x & 32 & 128 & 78.67 & 60.3 \\
\hyperlink{cite.survey_2412_09613}{\textcolor{blue}{PVC / InternVL2 8B}}\nocite{survey_2412_09613} & 64 & 16x & 96 & - & - & 64.1 \\
\hyperlink{cite.islam2025bimba}{\textcolor{blue}{BIMBA 7B}}\nocite{islam2025bimba} & 144 & 4x & 128 & 128 & 78.67 & - \\
\hyperlink{cite.zohar2024apollo}{\textcolor{blue}{Apollo 7B}}\nocite{zohar2024apollo} & 32 & 22.78x & 600 & - & - & 61.3 \\
\hyperlink{cite.survey_2505_15529}{\textcolor{blue}{Clapper 7B}}\nocite{survey_2505_15529} & 61.25 & 12.8x & 96 & 96 & 84.00 & 62.0 \\
\hyperlink{cite.survey_2412_09530}{\textcolor{blue}{Dynamic-VLM 7B}}\nocite{survey_2412_09530} & 100 & 5.76x & 256 & 120 & 80.00 & 60.9 \\
\hyperlink{cite.survey_2412_04729}{\textcolor{blue}{Espresso-align 7B}}\nocite{survey_2412_04729} & 0.5 at F=128 & 2P & 128 & 128 & 78.67 & - \\
\hyperlink{cite.wang2025internvideo2}{\textcolor{blue}{InternVideo2.5 7B}}\nocite{wang2025internvideo2} & 16 & - & 512 & - & - & 65.1 \\
\hyperlink{cite.survey_2503_13956}{\textcolor{blue}{F-16 7B}}\nocite{survey_2503_13956} & $\lfloor\sqrt{P}/2\rfloor^2/16$ & formula & 1760 & $\leq$1760 & capacity0 & - \\
\hyperlink{cite.survey_2409_12961}{\textcolor{blue}{Oryx 7B}}\nocite{survey_2409_12961} & N/4 or N/16 & 4x or16x & 64 /256 & - & - & 58.3 \\
\hyperlink{cite.survey_2412_05185}{\textcolor{blue}{LinVT / Qwen2-VL7B}}\nocite{survey_2412_05185} & - & - & - & - & - & 66.4 \\
\hyperlink{cite.survey_2504_20384}{\textcolor{blue}{FiLA-Video 7B}}\nocite{survey_2504_20384} & 32N/F & 3x temporal & 96 & - & - & 62.3 \\
\hyperlink{cite.survey_2406_09418}{\textcolor{blue}{VideoGPT +  3.8B}}\nocite{survey_2406_09418} & 208 & 4x dual & 16 & 16 & 97.33 & - \\
\multicolumn{7}{l}{\textbf{Reference video models}}\\*
\hyperlink{cite.zhang2024long}{\textcolor{blue}{LongVA 7B}}\nocite{zhang2024long} & 144 & 4x & N/A video & 128 & 78.67 & 52.6 \\
\hyperlink{cite.chen2024longvila}{\textcolor{blue}{LongVILA 7B-256f}}\nocite{chen2024longvila} & 256 & 4x & 256 & 256 & 57.33 & 60.1 \\
\hyperlink{cite.chen2024longvila}{\textcolor{blue}{LongVILA 7B-1M}}\nocite{chen2024longvila} & 196 & 3.72x & 2048 & - VMME & - & - \\
\hyperlink{cite.shen2025long}{\textcolor{blue}{Long-VITA 14B-128K}}\nocite{shen2025long} & 256 & 4x & 512 & 256 & 57.33 & 66.4 \\
\hyperlink{cite.liu2024nvila}{\textcolor{blue}{NVILA 8B-Video}}\nocite{liu2024nvila} & 32 & 32x & 256 & 256 & 57.33 & 64.2 \\
\hyperlink{cite.survey_minicpm_v26}{\textcolor{blue}{MiniCPM-V 2.6}}\nocite{survey_minicpm_v26} & 64 & 16x square* & - & - & - & 60.9 \\
\hyperlink{cite.li2024llava}{\textcolor{blue}{LLaVA-OneVision 7B}}\nocite{li2024llava} & 196 & 3.72x & 32 & 32 & 94.67 & 58.2 \\
\hyperlink{cite.zhang2024video}{\textcolor{blue}{LLaVA-Video 7B}}\nocite{zhang2024video} & 169 & 4.31x & 64 & 64 & 89.33 & 63.3 \\
\hyperlink{cite.bai2025qwen2}{\textcolor{blue}{Qwen2.5-VL 7B}}\nocite{bai2025qwen2} & Dynamic & 4x ViT & - & $\leq$768 & capacity0 & 65.1 \\
\multicolumn{7}{l}{\textbf{Inference-time methods (training maximum is N/A for the method)}}\\*
\hyperlink{cite.survey_2411_15024}{\textcolor{blue}{DyCoke  +  OV7B K=.5}}\nocite{survey_2411_15024} & 122.5 & 1.6x pooled & N/A & 32 & 94.67 & 59.5 \\
\hyperlink{cite.survey_2412_20504}{\textcolor{blue}{ReTaKe  +  Qwen2-VL7B}}\nocite{survey_2412_20504} & Dynamic input/KV & Separate stages & N/A & $\leq$2048 & capacity0* & - \\
\hyperlink{cite.survey_2412_20504}{\textcolor{blue}{ReTaKe  +  LLaVA-Video7B}}\nocite{survey_2412_20504} & Dynamic input/KV & Separate stages & N/A & $\leq$1024 & capacity0* & - \\
\hyperlink{cite.survey_2411_14401}{\textcolor{blue}{DyTo  +  LLaVA7B}}\nocite{survey_2411_14401} & 36.8 & 15.65x & N/A & 100 candidates & 83.33 + adaptive & 41.2 \\
\hyperlink{cite.survey_2506_02850}{\textcolor{blue}{METok  +  OV7B}}\nocite{survey_2506_02850} & Adaptive & Layer-dependent & N/A & 32 & 94.67 + adaptive & 58.4 \\
\end{longtable}
\endgroup

\section{Limitations}
\label{app:limitations}

This paper explores efficient multimodal large models for video understanding. Specifically, we train a video comprehension model from scratch using moderately-sized datasets based on a language model architecture. Benefiting from a dynamic compression-based efficient encoding scheme, the model processes video frames with high receptive fields during training and demonstrates competitive performance across multiple benchmarks during inference. While comprehensive ablation studies validate the model's effectiveness, two key limitations persist: (1) Training data quality - we observe that most pretraining video captions are summary-type texts, which hinders the model's ability to learn detailed video content understanding;
(2) Spatial complexity - we find that using the final layer features of ViT as input to the cubing network for video partitioning yields the best results, but this may not be the optimal approach in terms of spatial complexity. Future work will explore more efficient spatial complexity approaches, such as utilizing features from earlier layers of ViT combined with a more efficient cubing network structure for video partitioning.
We will also explore more efficient training data construction methods to enhance the model's learning of detailed video content understanding, such as utilizing video captioning models to generate more fine-grained captions for pretraining videos, and incorporating more video-text datasets with detailed annotations into the training data.

\section{Additional Evaluations}

\begin{table}[h]
    \centering
\begin{adjustbox}{width=\linewidth,center}
\renewcommand{\arraystretch}{1.2}
\setlength{\tabcolsep}{4mm}
\begin{tabular}{lccccccc}
\toprule  
\multirow{2}{*}{\textbf{Model}} & \textbf{NExT-QA}  & \textbf{ActivityNet-QA} & \multicolumn{5}{c}{\centering \textbf{Video-ChatGPT} } \\ \cline{4-8}
  & \cellcolor{gray!10} acc & \cellcolor{gray!10} acc & \cellcolor{gray!10} Correctness & \cellcolor{gray!10} Detail & \cellcolor{gray!10} Context & \cellcolor{gray!10} Temporal & \cellcolor{gray!10} Consistency \\
\midrule
  LLaMA-VID (7B)   & - & 47.4/3.3 & 2.96 & 3.00 & 3.53 & 2.46 & 2.51 \\
  Chat-UniVi (7B)   & - & 46.1/3.3 & 2.89 & 2.91 & 3.46 & 2.89 & 2.81 \\
  Video-LLaVA (7B)   & - & 45.3/3.3 & 2.87 & 2.94 & 3.44 & 2.45 & 2.51 \\
  VideoChat2 (7B)  &   68.6 & 49.1/3.3 & 3.16 & 3.08 & 3.69 & 2.56 & 3.14 \\
  VideoLLaMA2 (7B)  & 75.6 & 50.2/3.3 & 3.30 & 33.18 & 3.78 & 2.66 & 3.12 \\
  LLaVA-NeXT-Video (7B) &   78.2 & 53.5/3.2 & 3.39 & 3.29 & 3.92 & 2.60 & 3.12 \\
 \rowcolor{blue!10} Quicksviewer (8B)  & 77.5 & 47.6/2.7 & 3.10 & 3.11 & 3.09 & 2.48 & 3.04 \\
\bottomrule
\end{tabular}
\end{adjustbox}
\vspace*{2pt}
\caption{Evaluation results on more benchmarks.}
\label{tab:app-additional-evaluations}
\end{table}

\section{Qualitative Analysis}
\label{app:qualitative_analysis}
    
Our unified perception paradigm demonstrates efficient visual understanding capabilities, successfully processing: 1) lengthy videos, 2) high-resolution single images, and 3) multi-image contextual reasoning tasks. Through representative qualitative analysis, we validate the model's performance across these diverse inpus.
Specifically, we evaluate the video understanding capabilities of our model through documentary recording and sports competitions, while the image understanding proficiency across diverse domains including outdoor road scenes, physics/biological/historical knowledge systems, and multi-image geographical analysis.

In the documentary depicting a penguin chick's perilous encounter and subsequent escape, our model demonstrates comprehensive video understanding by: (1) identifying the nature of the unexpected attack, (2) precisely locating its temporal occurrence, and (3) summarizing the eventual outcome - showcasing its advanced capabilities in long-form video understanding, including temporal action recognition, event narrative abstraction, and exact timestamp localization.

In a lacrosse match video documenting a scoring play, our model precisely identifies the initiating player when queried about the offensive sequence, subsequently describing the play development and correctly specifying both the scoring player's identity and jersey number. This demonstrates the model's dual capability of (1) recognizing individual athletes in sports footage and (2) logically summarizing dynamic game situations.

\begin{figure*}[t!]
    \includegraphics[width=\textwidth]{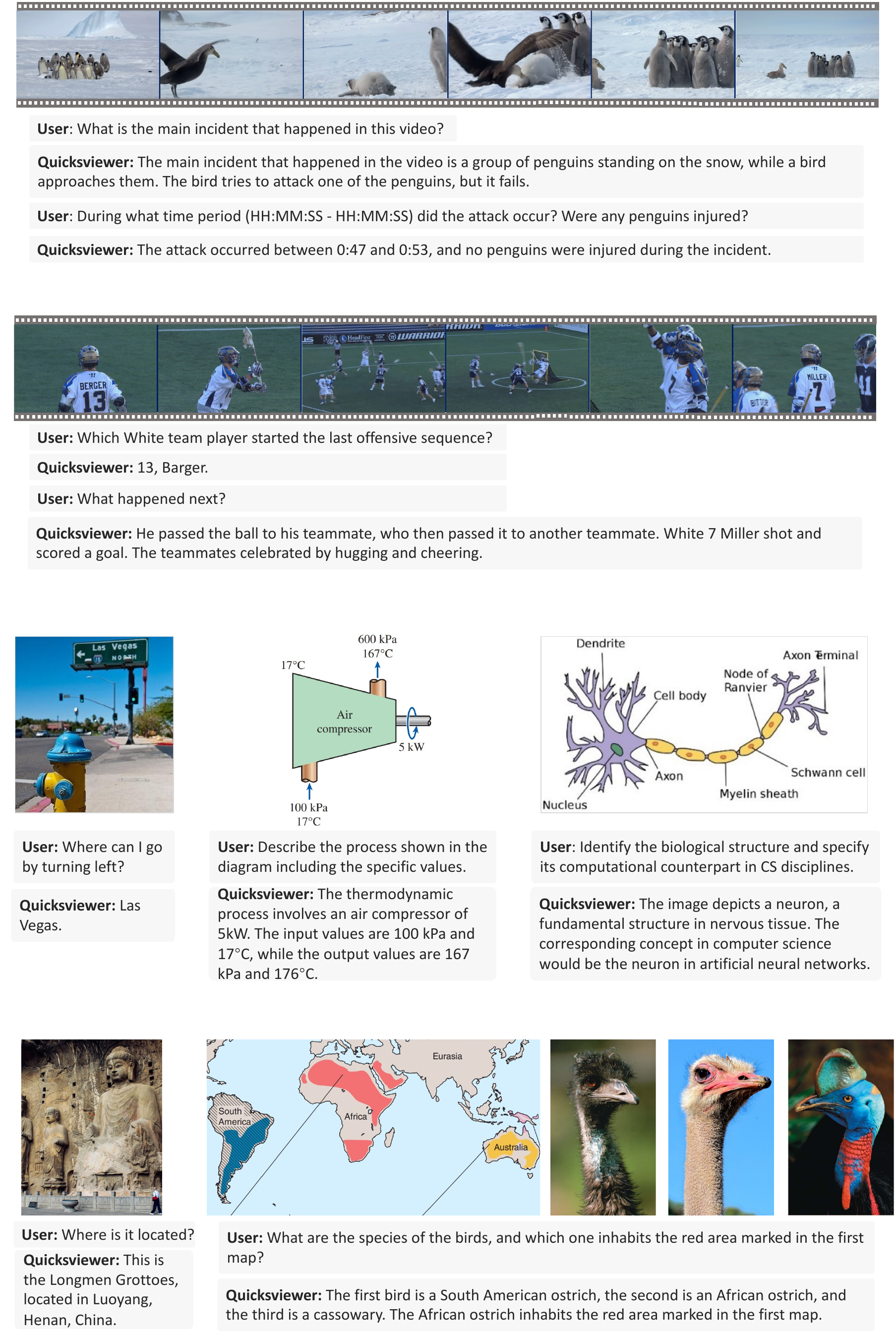}
    \caption{Qualitative analysis showns that Quicksviewer effectively understands lengthy documentary and sports videos, as well as informative single and multiple images.}
    \label{fig:main_cases}
\end{figure*}

We further validate our model's image understanding capabilities across extensive scenarios. As illustrated in Figure~\ref{fig:main_cases}, these includes: 1) traffic sign recognition in driving environments, 2) physics problem solving, 3) biological image interpretation and association, 4) historical scene identification, and 5) multi-image geographical reasoning. These examples demonstrate that while exhibiting strong video understanding, our model maintains robust image understanding capabilities. Benefiting from interleaved image-text training data, the model additionally acquires in-context learning capabilities for complex reasoning tasks.

\end{document}